\documentclass{article} % For LaTeX2e
\usepackage[accepted]{icml2018}
\usepackage{amsmath,amsfonts,bm}

\def\Figref#1{Fig.~\ref{#1}}

\def\Secref#1{Section~\ref{#1}}
\def\eqref#1{equation~\ref{#1}}
\def\1{\bm{1}}

\DeclareMathAlphabet{\mathsfit}{\encodingdefault}{\sfdefault}{m}{sl}
\SetMathAlphabet{\mathsfit}{bold}{\encodingdefault}{\sfdefault}{bx}{n}

\usepackage{url}
\usepackage{subfigure}

\usepackage{tikz}
\usetikzlibrary{arrows,shapes}
\usepackage{wrapfig}
\usepackage{xr-hyper}
\usepackage{hyperref}
\usepackage{siunitx}
\usepackage{float}

\usepackage [english]{babel}
\usepackage [autostyle, english = american]{csquotes}
\usepackage[subtle,title=normal,sections=normal,margins=normal,indent=normal,bibliography=normal]{savetrees} 
\MakeOuterQuote{"}

\usepackage{graphicx,wrapfig}

\newcommand{\ie}{i.e.}  % \emph{i.e.}}
\newcommand{\CBN}{CBN}
\newcommand{\CBNs}{CBNs}

\tikzstyle{dgraph}=[->, line width=1.5pt]
\tikzstyle{hcont}=[circle,draw=black!50,thick,fill=black!10,minimum size=6mm,>=stealth]  % hidden continuous node
\tikzstyle{ocont}=[circle,draw=blue!50,thick,fill=blue!10,minimum size=6mm,>=stealth]  % observed continuous node
\tikzstyle{contdec}=[circle,draw=red!50,thick,fill=red!10]  % observed node after a decision

\tikzstyle{cont}=[circle,draw=blue!50,thick,minimum size=6mm,line width=2pt,>=stealth]  % continuous  node
\tikzstyle{blackcont}=[circle,draw=black!50,thick,minimum size=6mm,line width=2pt,>=stealth]  % continuous  node
\tikzstyle{oval}=[ellipse,draw=blue!50,thick,minimum size=6mm,line width=2pt,>=stealth]  % continuous node
\tikzstyle{disc}=[rectangle,draw=blue!50,thick,line width=1pt,minimum size=6mm]  % discrete node

\tikzstyle{fillred}=[fill=red!20,thick]  % observed node
\tikzstyle{fillgreen}=[fill=green!20,thick]  % observed node
\tikzstyle{purered}=[fill=red]  % observed node
\tikzstyle{state}=[rectangle,fill=red!20]  % state
\tikzstyle{sobs}=[fill=green!15,thick]  % sequentally observed node
\tikzstyle{fact}=[fill,minimum size=1.5mm,line width=2pt,>=stealth]
\tikzstyle{varfact}=[draw,minimum size=1.5mm,line width=2pt,>=stealth]
\tikzstyle{sep}=[rectangle,draw=magenta!50,thick,minimum size=6mm]  % discrete node
\tikzstyle{det}=[fill=red!15,rectangle,draw=red!50,thick,minimum size=6mm]  % deterministic node

\tikzstyle{dethid}=[diamond,draw=red!50,thick,minimum size=6mm]  % deterministic  hidden node

\tikzstyle{lineball}=[fill,-*,draw=red!50,line width=1.5pt]
\tikzstyle{redball}=[mark=*,mark options={fill=red!50,draw=red},mark size=0.5pt]
\tikzstyle{greenball}=[mark=*,mark options={fill=green!50,draw=green},mark size=0.5pt]

\tikzstyle{dec}=[rectangle,draw=red!50,thick,minimum size=6mm]  % decision node
\tikzstyle{utility}=[diamond,draw=red!50,thick,minimum size=6mm]  % utility node
\tikzstyle{decutility}=[diamond,draw=red!50,thick,minimum size=6mm]  % utility node

\tikzstyle{contobs}+=[cont]
\tikzstyle{contobs}+=[obs]
\tikzstyle{discobs}+=[disc]
\tikzstyle{discobs}+=[obs]

\tikzstyle{obsred}+=[obs]
\tikzstyle{obsred}+=[red]

\tikzstyle{background grid}=[draw, black!50,step=.1cm]
\tikzstyle{dgraph}=[->, line width=1.5pt]
\tikzstyle{ugraph}=[line width=1.5pt]

\icmltitlerunning{Causal Reasoning from Meta-reinforcement Learning}

\begin{document}

\title{Causal Reasoning from Meta-reinforcement Learning}
\author{{\large \bf Ishita Dasgupta\thanks{Corresponding author: ishitadasgupta@g.harvard.edu}$^{~~1, 4}$ , Jane Wang$^{1}$,} \\
        {\large \bf Silvia Chiappa$^{1}$, Jovana Mitrovic$^{1}$, Pedro Ortega$^{1}$,} \\
        {\large \bf David Raposo$^{1}$, Edward Hughes$^{1}$, Peter Battaglia$^{1}$,} \\
        {\large \bf Matthew Botvinick$^{1,3}$, Zeb Kurth-Nelson$^{1,2}$} \\
\vspace{0.05 cm}\\
{\normalsize 
$^1$DeepMind, UK }\\
{\normalsize $^2$MPS-UCL Centre for Computational Psychiatry, UCL, UK }\\
{\normalsize $^3$Gatsby Computational Neuroscience Unit, UCL, UK }\\
{\normalsize $^4$Department of Physics and Center for Brain Science, Harvard University, USA} }
%\scriptsize\texttt{\{wangjane, csilvia, mitrovic, pedroortega, draposo, edwardhughes, peterbattaglia, botvinick, zebk\} @google.com} \\

\date{}
\maketitle

\begin{abstract}
Discovering and exploiting the causal structure in the environment is a crucial challenge for intelligent agents. Here we explore whether causal reasoning can emerge via meta-reinforcement learning. We train a recurrent network with model-free reinforcement learning to solve a range of problems that each contain causal structure. We find that the trained agent can perform causal reasoning in novel situations in order to obtain rewards. The agent can select informative interventions, draw causal inferences from observational data, and make counterfactual predictions. Although established formal causal reasoning algorithms also exist, in this paper we show that such reasoning can arise from model-free reinforcement learning, and suggest that causal reasoning in complex settings may benefit from the more end-to-end learning-based approaches presented here. This work also offers new strategies for structured exploration in reinforcement learning, by providing agents with the ability to perform---and interpret---experiments.
\end{abstract}

\section{Introduction}

Many machine learning algorithms are rooted in discovering patterns of correlation in data. While this has been sufficient to excel in several areas \citep{krizhevsky2012imagenet,cho2014learning}, sometimes the problems we are interested in are intrinsically causal. Answering questions such as "Does smoking cause cancer?" or "Was this person denied a job due to racial discrimination?" or "Did this marketing campaign cause sales to go up?" require an ability to reason about causes and effects. Causal reasoning may be an essential component of natural intelligence and is present in human babies, rats, and even birds \citep{leslie1982perception,gopnik2001causal,gopnik2004theory,blaisdell2006causal, lagnado2013causal}. 

There is a rich literature on formal approaches for defining and performing causal reasoning \citep{pearl2000,spirtes2000causation,dawid07fundamentals,pearl16causal}.
% This literature commonly decouples the causal reasoning task into the problem of causal induction (inferring the structure), and the problem of causal inference (estimating the effects of causes on the induced structure) and address each separately. In this paper, we develop a procedure where both steps of causal induction and inference can be performed in a single end-to-end procedure.
We investigate whether such reasoning can be achieved by meta-learning. The approach of meta-learning is to learn the learning (or inference/estimation) procedure itself, directly from data. Analogous \cite{grant2018recasting} models that learn causal structure directly from the environment, rather than having a pre-conceived formal theory, have also been implicated in human intelligence \cite{goodman2011learning}. 

We specifically adopt the "meta-reinforcement learning" method introduced previously \citep{duan2016RL2,wang2016}, in which a recurrent neural network (RNN)-based agent is trained with model-free reinforcement learning (RL). Through training on a large family of structured tasks, the RNN becomes a learning algorithm which generalizes to new tasks drawn from a similar distribution. In our case, we train on a distribution of tasks that are each underpinned by a different causal structure. We focus on abstract tasks that best isolate the question of interest: whether meta-learning can produce an agent capable of causal reasoning, when no notion of causality is explicitly given to the agent.

Meta-learning offers advantages of scalability by amortizing computations and, by learning end-to-end, the algorithm has the potential to find the internal representations of causal structure best suited for the types of causal inference required \citep{andrychowicz2016learning,wang2016,finn2017model}. We chose to focus on the RL approach because we are interested in agents that can learn about causes and effects not only from passive observations but also from active interactions with the environment \cite{hyttinen2013experiment,shanmugam2015learning}.

\section{Problem Specification and Approach}
We examine three distinct data settings -- \textit{observational}, \textit{interventional}, and \textit{counterfactual} -- which test different kinds of reasoning. 
\begin{itemize}%[leftmargin=*]
\setlength{\itemsep}{-2pt}  
\setlength{\parsep}{-4pt}
\vspace{-10pt}
\item In the observational setting~(Experiment~1), the agent can only obtain passive observations from the environment. This type of data allows an agent to infer correlations (\emph{associative reasoning}) and, depending on the structure of the environment, causal effects (\emph{cause-effect reasoning}). 
\item In the interventional setting~(Experiment~2), the agent can act in the environment by setting the values of some variables and observing the consequences on other variables. This type of data facilitates estimating causal effects.
\item In the counterfactual setting~(Experiment~3), the agent first has an opportunity to learn about the causal structure of the environment through interventions. At the last step of the episode, it must answer a counterfactual question of the form ``What \textit{would have} happened if a different intervention had been made in the previous timestep?''.
\vspace{-10pt}
\end{itemize}

Next we formalize these three settings and the patterns of reasoning possible in each, using the graphical model framework
\citep{pearl2000,spirtes2000causation,dawid07fundamentals}%\footnote{In this literature, the problem of causal induction (inferring the structure of the causal graph from data) is often decoupled from the problem of causal reasoning on the induced graph. In our experiments, however, the agent performs both steps in a single end-to-end procedure.}
.
Random variables will be denoted by capital letters (e.g., $E$) and their values by small letters (e.g., $e$). 

\subsection{Causal Reasoning\label{sec:causality}}
\begin{figure}[t]
\centering
\subfigure[]{
\scalebox{0.85}{
\begin{tikzpicture}[dgraph]
\node at (1,0.6) {${\cal G}$}; 
\node[ocont] (A) at (1,1.5) {$A$};
\node[ocont] (E) at (0,0) {$E$};
\node[ocont] (H) at (2,0) {$H$};
\node[] at (1,2.1) {$p(A)$};
\node[] at (0,-0.7) {$p(E|A)$};
\node[] at (2,-0.7) {$p(H|A,E)$};
\draw[line width=1.15pt](A)--(E);\draw[line width=1.15pt](A)--(H);\draw[line width=1.15pt](E)--(H);
\end{tikzpicture}}}
\hskip0.3cm
\subfigure[]{
\scalebox{0.8}{
\begin{tikzpicture}[dgraph]
\node at (0.9,0.6) {${\cal G}_{\rightarrow E=e}$}; 
\node[ocont] (A) at (1,1.5) {$A$};
\node[contdec] (E) at (0,0) {$E$};
\node[ocont] (H) at (2,0) {$H$};
\node[] at (1,2.1) {$p(A)$};
\node[] at (0,-0.7) {$\delta(E-e)$};
\node[] at (2,-0.7) {$p(H|A,E)$};
\draw[line width=1.15pt](A)--(H);\draw[line width=1.15pt](E)--(H);
\end{tikzpicture}}}
\vspace{-0.2cm}
\caption{(a): A \CBN~${\cal G}$ with a confounder for the effect of exercise ($E$) on heath ($H$) given by age ($A$). 
(b): Intervened \CBN~${\cal G}_{\rightarrow E=e}$ resulting from modifying ${\cal G}$ by replacing $p(E|A)$ with a delta distribution $\delta(E-e)$ and leaving the remaining conditional distributions $p(H|E,A)$ and $p(A)$ unaltered.}
\label{fig:CBN}
\vspace{-10pt}
\end{figure}
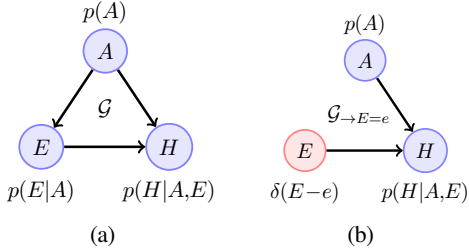

Causal relationships among random variables can be expressed using \emph{causal Bayesian networks} (\CBNs) (see the Supplementary Material). A \CBN~is a directed acyclic graphical model that captures both \emph{independence} and \emph{causal} relations. Each node $X_i$ corresponds to a random variable, and the joint distribution $p(X_1, \ldots, X_N)$ is given by the product of conditional distributions of each node $X_i$ given its parent nodes $\textrm{pa}(X_i)$, \ie~$p(X_{1:N}\equiv X_1,\ldots,X_N)=\prod_{i=1}^Np(X_i|\textrm{pa}(X_i))$. 

Edges carry causal semantics: %if a variable $X_i$ has an edge pointing at $X_j$, then $X_i$ is a \emph{direct cause} of $X_j$; 
if there exists a directed path from $X_i$ to $X_j$, then $X_i$ is a \emph{potential cause} of $X_j$. Directed paths are also called \emph{causal paths}. The \emph{causal effect} of $X_i$ on $X_j$ is the conditional distribution of $X_j$ given $X_i$ restricted to only causal paths.

An example of \CBN~${\cal G}$ is given in \Figref{fig:CBN}a, where $E$ represents hours of exercise in a week, $H$ cardiac health, and $A$ age.
The causal effect of $E$ on $H$ is the conditional distribution restricted to the path $E\rightarrow H$, \ie~excluding the path $E\leftarrow A \rightarrow H$. The variable $A$ is called a \emph{confounder}, as it confounds the causal effect with non-causal statistical influence. 
Simply observing cardiac health conditioning on exercise level from $p(H|E)$ (associative reasoning) cannot answer if change in exercise levels cause changes in cardiac health (cause-effect reasoning), since there is always the possibility that correlation between the two is because of the common confounder of age. %Another key point is that such correlations cannot answer if exercise causes changes in cardiac health or if cardiac health causes changes in exercise level.

\textbf{Cause-effect Reasoning.}
The causal effect of $E=e$ can be seen as the conditional distribution $p_{\rightarrow E=e}(H|E=e)$\footnote{In the causality literature, this distribution would most often be indicated with $p(H|\textrm{do}(E=e))$. We prefer to use $p_{\rightarrow E=e}(H|E=e)$ to highlight that intervening on $E$ results in changing the original distribution $p$, by structurally altering the \CBN.} on the \emph{intervened} \CBN~${\cal G}_{\rightarrow E=e}$ resulting from replacing $p(E|A)$ with a delta distribution $\delta(E-e)$ (thereby removing the link from $A$ to $E$) and leaving the remaining conditional distributions $p(H|E,A)$ and $p(A)$ unaltered (\Figref{fig:CBN}b).
The rules of do-calculus \citep{pearl2000,pearl16causal} tell us how to compute $p_{\rightarrow E=e}(H|E=e)$ using observations from ${\cal G}$. In this case $p_{\rightarrow E=e}(H|E=e) = \sum_A p(H|E=e,A)p(A)$\footnote{Notice that conditioning on $E = e$ would instead give $p(H|E=e)=\sum_{A}p(H|E=e,A)p(A|E=e)$.}.
Therefore, do-calculus enables us to reason in the intervened graph ${\cal G}_{\rightarrow E=e}$ even if our observations are from ${\cal G}$. This is the scenario captured by our observational data setting outlined above. 

Such inferences are always possible if the confounders are observed, but in the presence of unobserved confounders, for many \CBN~structures the only way to compute causal effects is by collecting observations directly from the intervened graph, e.g. from ${\cal G}_{\rightarrow E=e}$ by fixing the value of the variable $E=e$ and observing the remaining variables---we call this process performing an actual intervention in the environment. In our interventional data setting, outlined above, the agent has access to such interventions. 

\textbf{Counterfactual Reasoning.} Cause-effect reasoning can be used to correctly answer predictive questions of the type "Does exercising improve cardiac health?" by accounting for causal structure and confounding. However, it cannot answer retrospective questions about what \textit{would have} happened. For example, given an individual $i$ who has died of a heart attack, this method would not be able to answer questions of the type "What would the cardiac health of this individual have been had she done more exercise?". This type of question requires
reasoning about a counterfactual world (that did not happen). To do this, we can first use the observations from the factual world and knowledge about the \CBN~to get an estimate of the specific latent randomness in the makeup of individual $i$ (for example information about this specific patient's blood pressure and other variables as inferred by her having had a heart attack). Then, we can use this estimate to compute cardiac health under intervention on exercise. This procedure is explained in more detail in the Supplementary Material.
%augment our prediction for the effects of intervening on exercise $E$ in that world. This procedure is explained in more detail in the Supplementary Material. 

\subsection{Meta-learning}\label{sec:metal}

Meta-learning refers to a broad range of approaches in which aspects of the learning algorithm itself are learned from the data. Many individual components of deep learning algorithms have been successfully meta-learned, including the optimizer \citep{andrychowicz2016learning}, initial parameter settings \citep{finn2017model}, a metric space \citep{vinyals2016matching}, and use of external memory \citep{santoro2016meta}. 

Following the approach of \citep{duan2016RL2,wang2016}, we parameterize the entire learning algorithm as a recurrent neural network (RNN), and we train the weights of the RNN with model-free reinforcement learning (RL). The RNN is trained on a broad distribution of problems which each require learning. When trained in this way, the RNN is able to implement a learning algorithm capable of efficiently solving novel learning problems in or near the training distribution (see the Supplementary Material for further details). 

Learning the weights of the RNN by model-free RL can be thought of as the "outer loop" of learning. The outer loop shapes the weights of the RNN into an "inner loop" learning algorithm. This inner loop algorithm plays out in the activation dynamics of the RNN and can continue learning even when the weights of the network are frozen. The inner loop algorithm can also have very different properties from the outer loop algorithm used to train it. For example, in previous work this approach was used to negotiate the exploration-exploitation tradeoff in multi-armed bandits \citep{duan2016RL2} and learn algorithms which dynamically adjust their own learning rates \citep{wang2016,wang2018}. In the present work we explore the possibility of obtaining a causally-aware inner-loop learning algorithm.

\section{Task Setup and Agent Architecture}
\label{sec:task}

In our experiments, in each episode the agent interacted with a different \CBN~${\cal G}$, defined over a set of $N$ variables. The structure of ${\cal G}$ was drawn randomly from the space of possible acyclic graphs under the constraints given in the next subsection.

Each episode consisted of $T$ steps, which were divided into two phases: an \emph{information phase} and a \emph{quiz phase}. The information phase, corresponding to the first $T-1$ steps, allowed the agent to collect information by interacting with or passively observing samples from ${\cal G}$. The agent could potentially use this information to infer the connectivity and weights of ${\cal G}$. The quiz phase, corresponding to the final step $T$, required the agent to exploit the causal knowledge it collected in the information phase, to select the node with the highest value under a random external intervention.

\subsection*{Causal Graphs, Observations, and Actions}

We generated all graphs on $N=5$ nodes, with edges only in the upper triangular of the adjacency matrix (this guarantees that all the graphs obtained are acyclic). Edge weights $w_{ji}$ were uniformly sampled from $\{-1, 0, 1\}$. This yielded $3^{N(N - 1)/2}=59049$ unique graphs. These can be divided into equivalence classes: sets of graphs that are structurally identical but differ in the permutation of the node labels. Our held-out test set consisted of $12$ random graphs plus all other graphs in the equivalence classes of these 12. Thus, all graphs in the test set had never been seen (and no equivalent graphs had been seen) during training. There were 408 total graphs in the test set.

Each node, $X_i \in \mathbb{R}$, was a Gaussian random variable. Parentless nodes had distribution $\mathcal{N}(\mu=0.0, \sigma=0.1)$. A node $X_i$ with parents $\textrm{pa}(X_{i})$ had conditional distribution $p(X_i \vert \textrm{pa}(X_{i})) = \mathcal{N}(\mu=\sum_{j} w_{ji}X_{j}, \sigma=0.1)$, where $X_j \in \textrm{pa}(X_{i})$\footnote{We also tested graphs with non-linear causal effects, and larger graphs of size N = 6 (see the Supplementary Material).}.

A root node of ${\cal G}$ was always hidden (unobservable), to allow for the presence of an unobserved confounder and the agent could therefore only observe the values of the other 4 nodes. The concatenated values of the nodes, $v_t$, and and a one-hot vector indicating the external intervention during the quiz phase, $m_{t}$, (explained below) formed the observation vector provided to the agent at step $t$, $o_{t}=[v_{t}, m_{t}]$.

In both phases, on each step $t$, the agent could choose to take 1 of $2(N-1)$ actions, the first $N-1$ of which were \emph{information actions}, and the second of which were \emph{quiz actions}. Both information and quiz actions were associated with selecting the $N-1$ observable nodes, but could only be legally used in the appropriate phase of the task. If used in the wrong phase, a penalty was applied and the action produced no effect.

\textbf{Information Phase.}
In the information phase, an information action $a_t = i$ caused an intervention on the $i$-th node, setting the value of $X_{a_t} = X_i = 5$ (the value $5$ was chosen to be outside the likely range of sampled observations, to facilitate learning the causal graph). The node values $v_t$ were then obtained by sampling from $p_{\rightarrow X_i=5}(X_{1:N\setminus i }|X_i=5)$ (where $X_{1:N\setminus i }$ indicates the set of all nodes except $X_i$), namely from the intervened \CBN~${\cal G}_{\rightarrow X_{a_t}=5}$ resulting from removing the incoming edges to $X_{a_t}$ from ${\cal G}$, and using the intervened value $X_{a_t} = 5$ for conditioning its children's values. 
If a quiz action was chosen during the information phase, it was ignored; namely, the node values were sampled from ${\cal G}$ as if no intervention had been made, and the agent was given a penalty of $r_t = -10$ in order to encourage it to take quiz actions during the quiz phase. There was no other reward during the information phase.

The default length an episode for fixed to be $T=N=5$, resulting in this phase being fixed to a length of $T-1=4$. This was because in the noise-free limit, a minimum of $N-1 = 4$ interventions, one on each observable node, are required in general to resolve the causal structure and score perfectly on the test phase.

\textbf{Quiz Phase.}
In the quiz phase, one non-hidden node $X_j$ was selected at random to be intervened on by the environment. Its value was set to $-5$. We chose $-5$ to disallow the agent from memorizing the results of interventions in the information phase (which were fixed to $+5$) in order to perform well on the quiz phase. The agent was informed which node received this external intervention via the one-hot vector $m_t$ as part of the observation from the the final pre-quiz phase timestep, $T-1$. For steps $t < T-1$, $m_t$ was the zero vector. The agent's reward on this step was the sampled value of the node it selected during the quiz phase. In other words, $r_T = X_i = X_{a_T-(N-1)}$ if the action selected was a quiz action (otherwise, the agent was given a penalty of $r_T = -10$).

\textbf{Active vs Random Agents.}
Our agents had to perform two distinct tasks during the information phase: a) actively choose which nodes to set values on, and b) infer the \CBN~from its observations. We refer to this setup as the ``active'' condition.
To better understand the role of (a), we include comparisons with a baseline agent in the ``random'' condition, whose policy is to choose randomly which observable node it will set values for, at each step of the information phase.
% To control for (a), we created the ``passive'' condition, where the agent's information phase actions are not learned. To provide a benchmark for how well the active agent can perform task (a), we fixed the passive agent's intervention policy to be an exhaustive sweep through all observable nodes. This is close to optimal for this domain -- in fact it is the optimal policy for noise-free conditional node values. We also compared the active agent's performance to a baseline agent whose policy is to intervene randomly on the observable nodes in the information phase, in the Supplementary Material.

\textbf{Two Kinds of Learning.}
The ``inner loop'' of learning (see \Secref{sec:metal}) occurs within each episode where the agent is learning from the evidence it gathers during the information phase in order to perform well in the quiz phase. The same agent then enters a new episode, where it has to repeat the task on a different \CBN. Test performance is reported on \CBNs~that the agent has never previously seen, after all the weights of the RNN have been fixed. Hence, the only transfer from training to test (or the ``outer loop'' of learning) is the ability to discover causal dependencies based on observations in the information phase, and to perform causal inference in the quiz phase.

\iffalse
\begin{figure*}[t]
\centering
  \subfloat[Description of one episode]{{\includegraphics[width=0.53\textwidth]{One_episode.png} }}%
    \qquad
    \subfloat[Training schema]{{\includegraphics[width=0.40\textwidth]{train_schema.png} }}%
    \caption{Methods. a) Structure of one episode, showing the inputs to the agent $[o_t, a_{t-1}, r_{t-1}]$ concatenated at each time step $t$ \comment{$T$ is the last time-step in the text. We need to change the figure. Is this figure really necessary?} \zeb{panel b is my stupid scribbling, but i kinda like panel a}, where $o_t = [v_t, m_t]$ concatenated, and it's output to the environment $[a_t]$. b) Outline of the meta-learning training framework, showing the inner loop of learning within an episode, and the outer loop of learning across episodes.}%
    \label{fig:methods}%
\end{figure*}
\fi

\subsection*{Agent Architecture and Training}

We used a long short-term memory (LSTM) network \citep{hochreiter97long} (with 192 hidden units) that, at each time-step $t$, receives a concatenated vector containing $[o_{t}, a_{t-1}, r_{t-1}]$ 
as input, where $o_{t}$ is the observation\footnote{'Observation' $o_{t}$ refers to the reinforcement learning term, \ie~the input from the environment to the agent. This is distinct from observations in the causal sense (referred to as observational data) \ie~samples from a causal structure where no interventions have been carried out.}, $a_{t-1}$ is the previous action (as a one-hot vector) and $r_{t-1}$ the reward  (as a single real-value)\footnote{These are both set to zero for the first step in an episode.}.

The outputs, calculated as linear projections of the LSTM's hidden state, are a set of policy logits (with dimensionality equal to the number of available actions), plus a scalar baseline. The policy logits are transformed by a softmax function, and then sampled to give a selected action. 

Learning was by \emph{asynchronous advantage actor-critic} \citep{mnih2016}. In this framework, the loss function consists of three terms -- the policy gradient, the baseline cost and an entropy cost. The baseline cost was weighted by 0.05 relative to the policy gradient cost. The weighting of the entropy cost was annealed over the course of training from 0.25 to 0. Optimization was done by RMSProp with $\epsilon = 10^{-5}$, momentum = 0.9 and decay = 0.95. Learning rate was annealed from $9\times 10^{-6}$ to 0, with a discount of 0.93. Unless otherwise stated, training was done for $1\times10^7$ steps using batched environments with a batch size of 1024.

For all experiments, after training, the agent was tested with the learning rate set to zero, on a held-out test set.

\section{Experiments}

Our three experiments (observational, interventional, and counterfactual data settings) differed in the properties of the $v_t$ that was observed by the agent during the information phase, and thereby limited the extent of causal reasoning possible within each data setting. Our measure of performance is the reward earned in the quiz phase for held-out \CBNs. Choosing a random node in the quiz phase results in a reward of $-5 / 4 = -1.25$, since one node (the externally intervened node) always has value $-5$ and the others have on average $0$ value. By learning to simply avoid the externally intervened node, the agent can earn on average $0$ reward. Consistently picking the node with the highest value in the quiz phase requires the agent to perform causal reasoning. For each agent, we took the average reward earned across 1632 episodes (408 held-out test \CBNs, with 4 possible external interventions). We trained 8 copies of each agent and reported the average reward earned by these, with error bars showing 95$\%$ confidence intervals. The $p$ values based on the appropriate t-test are provided in cases where the compared values are close.

\subsection{Experiment 1: Observational Setting}
\label{sec:expt1}

\begin{figure*}[!ht]
\centering
   \includegraphics[width=0.99\textwidth]{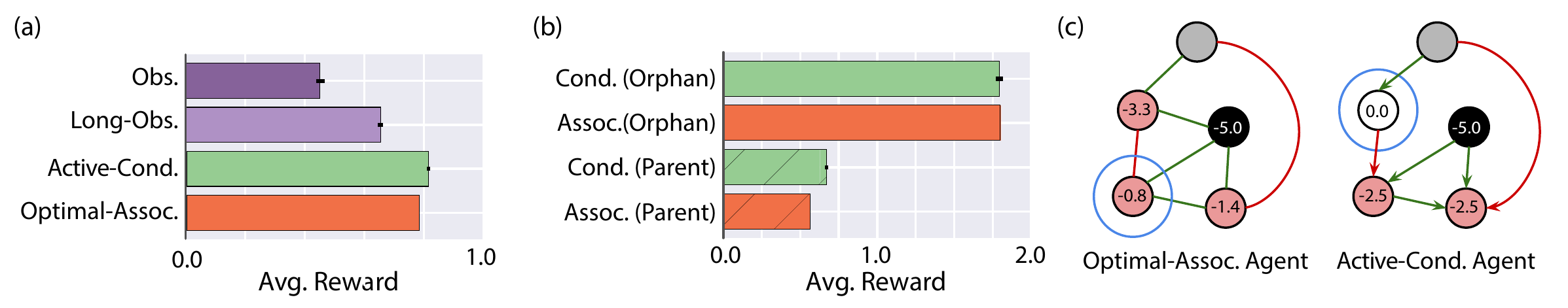}
%   \vspace{-0.3cm}
    \caption{Experiment 1. Agents do cause-effect reasoning from observational data. a) Average reward earned by the agents tested in this experiment. See main text for details. b) Performance split by the presence or absence of at least one parent (Parent and Orphan respectively) on the externally intervened node. c) Quiz phase for a test \CBN. Green (red) edges indicate a weight of $+1$ ($-1$). Black represents the intervened node, green (red) nodes indicate a positive (negative) value at that node, white indicates a zero value. The blue circles indicate the agent's choice. Left panel: The undirected version of ${\cal G}$ and the nodes taking the mean values prescribed by $p(X_{1:N\setminus j }|X_j=-5)$, including backward inference to the intervened node's parent. We see that the Optimal Associative Baseline's choice is consistent with maximizing these (incorrect) node values. Right panel: ${\cal G}_{\rightarrow X_j=-5}$ and the nodes taking the mean values prescribed by $p_{\rightarrow X_j=-5}(X_{1:N\setminus j }|X_j=-5)$. We see that the Active-Conditional Agent's choice is consistent with maximizing these (correct) node values.}%
    \vspace{-0.5cm}
    \label{fig:expt1}%
\end{figure*}

In Experiment~1, the agent could neither intervene to set the value of variables in the environment, nor observe any external interventions. In other words, it only received observations from ${\cal G}$, not ${\cal G}_{\rightarrow X_j}$ (where $X_j$ is a node that has been intervened on). This limits the extent of causal inference possible.
In this experiment, we tested five agents, four of which were learned: "Observational", "Long Observational", "Active Conditional", "Random Conditional", and the "Optimal Associative Baseline" (not learned). We also ran two other standard RL baselines---see the Supplementary Material for details.

\textbf{Observational Agents}: In the information phase, the actions of the agent were ignored\footnote{These agents also did not receive the out-of-phase action penalties during the information phase since their actions are totally ignored.}, and the observational agent always received the values of the observable nodes as sampled from the joint distribution associated with ${\cal G}$. In addition to the default $T=5$ episode length, we also trained this agent with 4$\times$ longer episode length (Long Observational Agent), to measure performance increase with more observational data.

\textbf{Conditional Agents}:
The information phase actions corresponded to observing a world in which the selected node $X_j$ is equal to $X_j=5$, and the remaining nodes are sampled from the conditional distribution \(p(X_{1:N\setminus j }|X_j=5)\).
This differs from intervening on the variable $X_{j}$ by setting it to the value $X_{j}=5$, since here we take a conditional sample from ${\cal G}$ rather than from ${\cal G}_{\rightarrow X_j=5}$ (or from \(p_{\rightarrow X_j=5}(X_{1:N\setminus j }|X_j=5)\)), and inference about the corresponding node's parents is possible. Therefore, this agent still has access to only observational data, as with the observational agents. However, on average it receives more diagnostic information about the relation between the random variables in ${\cal G}$, since it can observe samples where a node takes a value far outside the likely range of sampled observations.
We run active and random versions of this agent as described in \Secref{sec:task}.

\textbf{Optimal Associative Baseline:}
This baseline receives the true joint distribution $p(X_{1:N})$ implied by the \CBN~in that episode and therefore has full knowledge of the correlation structure of the environment\footnote{Notice that the agent does not know the graphical structure, \ie~it does not know which nodes are parents of which other nodes.}. It can therefore do exact associative reasoning of the form $p(X_j|X_i=x)$, but cannot do any cause-effect reasoning of the form $p_{\rightarrow X_i=x}(X_j|X_i=x)$. In the quiz phase, this baseline chooses the node that has the maximum value according to the true $p(X_j|X_i=x)$ in that episode, where $X_i$ is the node externally intervened upon, and $x = -5$. This is the best possible performance using only associative reasoning.

% \textbf{Observational MAP Baseline}: This baseline follows the traditional method of separating causal induction and causal inference. We first carry out exact maximum a posteriori (MAP) inference over the space of \CBNs~in each episode (\ie~causal induction) by selecting the \CBN~(${\cal G}^{\textrm{MAP}}$) of the $59049$ unique possibilities that maximizes the likelihood of the data observed, $v_{1:T}$, by the Observational Agent in that episode. This is equivalent to maximizing the posterior probability since the prior over graphs is uniform.

\begin{figure*}[t!]
\centering
   \includegraphics[width=0.99\textwidth]{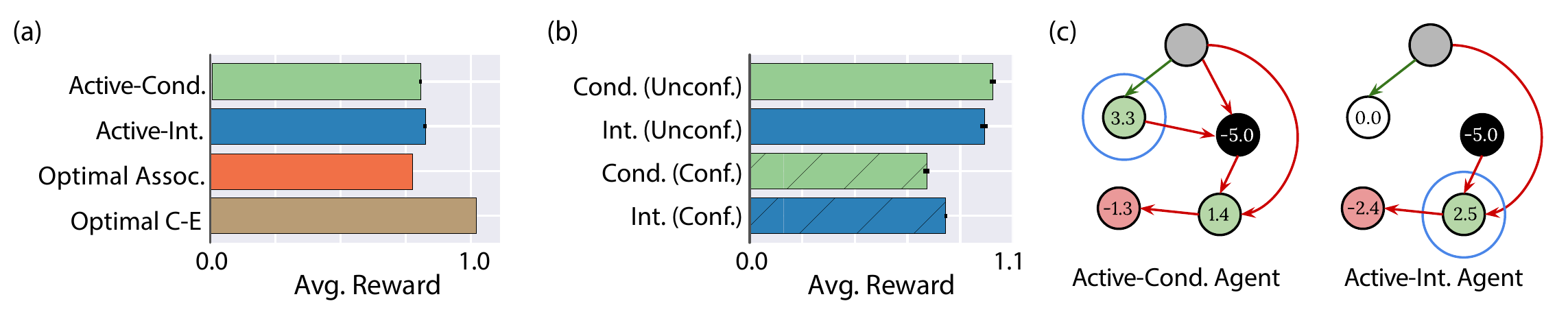}
%   \vspace{-0.3cm}
    \caption{Experiment 2. Agents do cause-effect reasoning from interventional data. a) Average reward earned by the agents tested in this experiment. See main text for details. b) Performance split by the presence or absence of unobserved confounders (abbreviated as Conf. and Unconf. respectively) on the externally intervened node. c) Quiz phase for a test \CBN. See \Figref{fig:expt1} for a legend. Here, the left panel shows the full $\cal G$ and the nodes taking the mean values prescribed by $p(X_{1:N\setminus j }|X_j=-5)$. %including an inc inference about a node that is confounded with the externally intervened node.
    We see that the Active-Cond Agent's choice is consistent with choosing based on these (incorrect) node values. The right panel shows ${\cal G}_{\rightarrow X_j=-5}$ and the nodes taking the mean values prescribed by $p_{\rightarrow X_j=-5}(X_{1:N\setminus j }|X_j=-5)$. We see that the Active-Int. Agent's choice is consistent with maximizing on these (correct) node value.}%
    \vspace{-0.5cm}
    \label{fig:expt2}%
\end{figure*}

\subsection*{Results}

We focus on two questions in this experiment.

(i) Most centrally, do the agents learn to perform cause-effect reasoning using observational data? The Optimal Associative Baseline tracks the greatest reward that can be achieved using only knowledge of correlations -- without causal knowledge. Compared to this baseline, the Active-Conditional Agent (which is allowed to select highly informative observations) earns significantly more reward ($p=$ \num{6e-5}, \Figref{fig:expt1}a). 
To better understand why the agent outperforms the associative baseline, we divided episodes according to whether or not the node that was intervened on in the quiz phase has a parent (\Figref{fig:expt1}b). If the intervened node $X_j$ has no parents, then ${\cal G}  = {\cal G}_{\rightarrow X_j}$, and cause-effect reasoning has no advantage over associative reasoning. Indeed, the Active-Conditional Agent performs better than the Optimal Associative Baseline only when the intervened node has parents (hatched bars in \Figref{fig:expt1}b). We also show the quiz phase for an example test CBN in \Figref{fig:expt1}c, where the Optimal Associative Baseline chooses according to the node values predicted by $\cal G$, whereas the Active-Conditional Agent chooses according the node values predicted by ${\cal G}_{\rightarrow X_j}$. 
\begin{wrapfigure}{l}{0.27\textwidth}
\centering
\vspace{-0.4cm}
\includegraphics[width=0.3\textwidth]{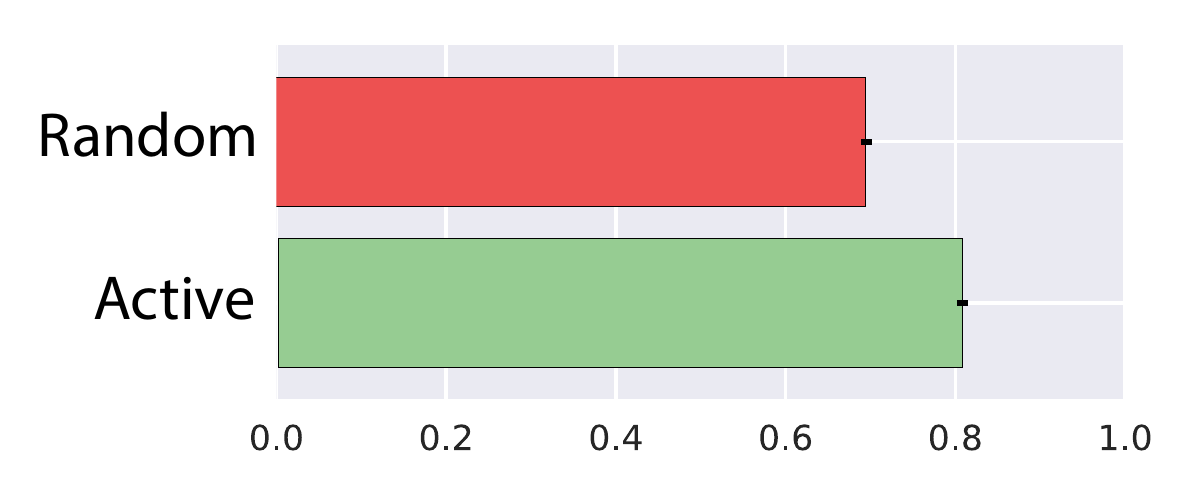}
\vspace{-20pt}
\caption{Active and Random Conditional Agents}
% \label{fig:expt1c}
\label{fig:expt1_active}
\vspace{-0.4cm}
\end{wrapfigure} 
These analyses allow us to conclude that the agent can perform cause-effect reasoning, using observational data alone -- analogous to the formal use of do-calculus.
(ii) Do the agents learn to select useful observations? We find that the Active-Conditional Agent's performance is significantly greater than the Random-Conditional Agent (\Figref{fig:expt1_active}). This indicates that the agent has learned to choose useful data to observe.

For completeness we also included agents that received unconditional observations from $\cal G$, i.e. the Observational Agents ('Obs' and 'Long-Obs' in \Figref{fig:expt1}a). As expected, these agents performed worse than the Active-Conditional Agent, because they received less diagnostic information during the information phase. However, they were still able to acquire some information from unconditional samples, and also made use of the increased information available from longer episodes.

\subsection{Experiment 2: Interventional Setting}
\label{sec:expt2}

In Experiment~2, the agent receives interventional data in the information phase -- it can choose to intervene on any observable node, $X_j$, and observe a sample from the resulting graph ${\cal G}_{\rightarrow X_j}$. As discussed in \Secref{sec:causality}, access to interventional data permits cause-effect reasoning even in the presence of unobserved confounders, a feat which is in general impossible with access only to observational data.
In this experiment, we test three new agents, two of which were learned: "Active Interventional", "Random Interventional", and "Optimal Cause-Effect Baseline" (not learned).

\textbf{Interventional Agents:}
The information phase actions correspond to performing an intervention on the selected node $X_j$ and sampling from ${\cal G}_{\rightarrow X_j}$ (see \Secref{sec:task} for details). We run active and random versions of this agent as described in \Secref{sec:task}.

% \textbf{Interventional MAP Baseline}: This baseline infers a CBN by maximizing the likelihood of the data observed by the Passive Interventional Agent in that episode.

% In the quiz phase, we predict the values of each node according to ${\cal G}_{\rightarrow X_j}^{\textrm{MAP}}$  where $X_j$ is the node externally intervened upon (\ie~causal inference), and choose the node with the highest value.

\textbf{Optimal Cause-Effect Baseline:}
This baseline receives the true \CBN, $\cal G$. In the quiz phase, it chooses the node that has the maximum value according to ${\cal G}_{\rightarrow X_j}$, where $X_j$ is the node externally intervened upon. This is the maximum possible performance on this task.

\subsection*{Results}

We focus on two questions in this experiment.

(i) Do our agents learn to perform cause-effect reasoning from interventional data? The Active-Interventional Agent's performance is marginally better than the Active-Conditional Agent ($p=0.06$, \Figref{fig:expt2}a). 
%\ishita{Is not obvious from graph -- can we add a significance test here?}. 
To better highlight the crucial role of interventional data in doing cause-effect reasoning, we compare the agent performances split by whether the node that was intervened on in the quiz phase of the episode had unobserved confounders with other variables in the graph (\Figref{fig:expt2}b). In confounded cases, as described in \Secref{sec:causality}, cause-effect reasoning is impossible with only observational data. We see that the performance of the Active-Interventional Agent is significantly higher ($p=10^{-5}$) than that of the Active-Conditional Agent in the confounded cases. This indicates that the Active-Interventional Agent (that had access to interventional data) is able to perform additional cause-effect reasoning in the presence of confounders that the Active-Conditional Agent (that had access to only observational data) cannot do.
This is highlighted by \Figref{fig:expt2}c, which shows the quiz phase for an example \CBN, where the Active-Conditional Agent is unable to resolve the unobserved confounder, but the Active-Interventional Agent is able to.
\begin{wrapfigure}{l}{0.27\textwidth}
\centering
\vspace{-0.4cm}
\includegraphics[width=0.3\textwidth]{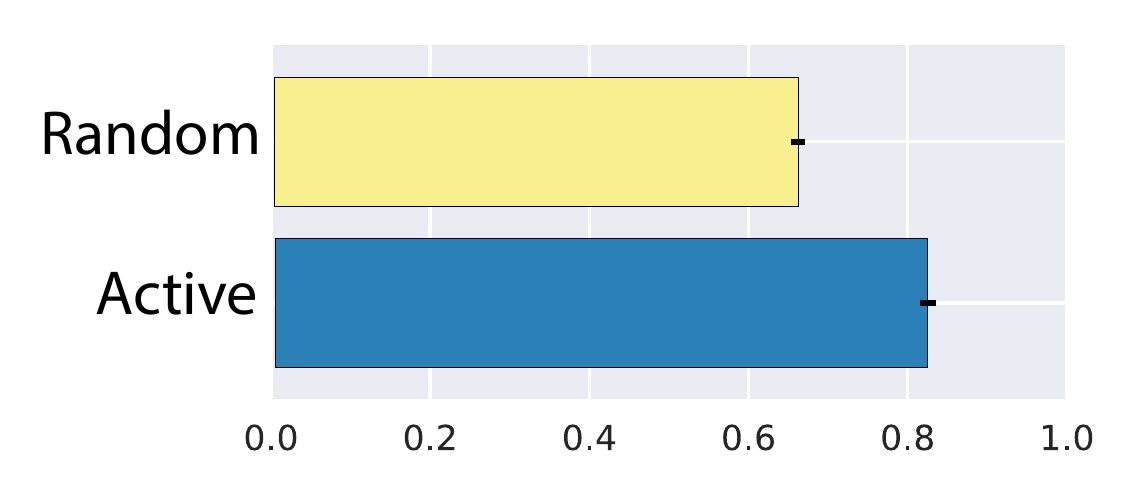} 
\vspace{-20pt}
\caption{Active and Random Interventional Agents}
\label{fig:expt2_active}
\vspace{-0.4cm}
\end{wrapfigure}(ii) Do our agents learn to make useful interventions? The Active-Interventional Agent's performance is significantly greater than the Random-Interventional Agent's (\Figref{fig:expt2_active}). This indicates that when the agent is allowed to choose its actions, it makes tailored, non-random choices about the interventions it makes and the data it wants to observe.

\subsection{Experiment 3: Counterfactual Setting}
\label{sec:expt3}

\begin{figure*}[t!]
\centering
   \includegraphics[width=0.95\textwidth]{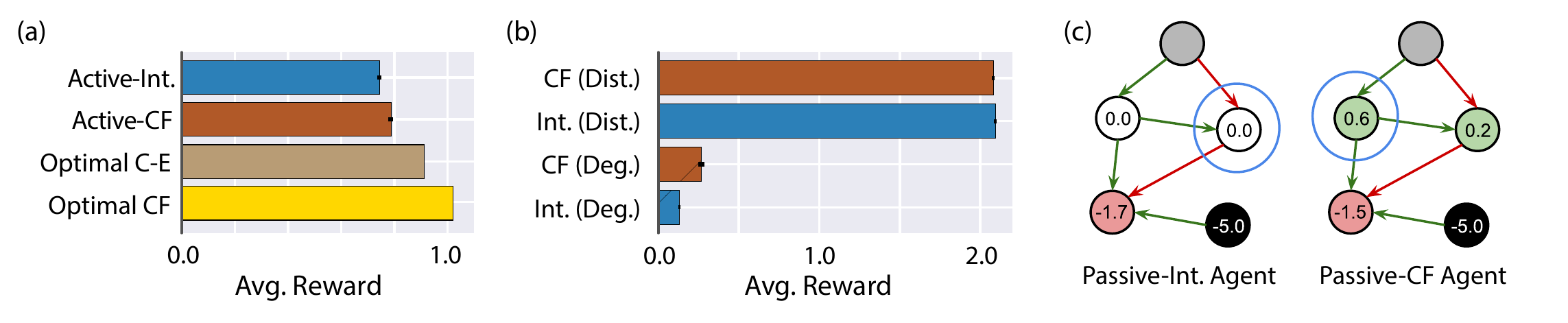}
%   \vspace{-0.3cm}
    \caption{Experiment 3.  Agents do counterfactual reasoning. a) Average reward earned by the agents tested in this experiment. See main text for details. b) Performance split by if the maximum node value in the quiz phase is degenerate (Deg.) or distinct (Dist.). c) Quiz phase for an example test-CBN. See \Figref{fig:expt1} for a legend. Here, the left panel shows ${\cal G}_{\rightarrow X_j=-5}$ and the nodes taking the mean values prescribed by $p_{\rightarrow X_j=-5}(X_{1:N\setminus j }|X_j=-5)$. We see that the Active-Int. Agent's choice is consistent with maximizing on these node values, where it makes a random choice between two nodes with the same value. The right panel panel shows ${\cal G}_{\rightarrow X_j=-5}$ and the nodes taking the exact values prescribed by the means of $p_{\rightarrow X_j=-5}(X_{1:N\setminus j }|X_j=-5)$, combined with the specific randomness inferred from the previous time step. As a result of accounting for the randomness, the two previously degenerate maximum values are now distinct. We see that the Active-CF. agent's choice is consistent with maximizing on these node values.}%
    \vspace{-0.5cm}
    \label{fig:expt3}%
\end{figure*}

In Experiment~3, the agent was again allowed to make interventions as in Experiment~2, but in this case the quiz phase task entailed answering a counterfactual question. We explain here what a counterfactual question in this domain looks like. Assume $X_i = \sum_{j} w_{ji}X_{j} + \epsilon_i$ where $\epsilon_i$ is distributed as $\mathcal{N}(0.0, 0.1)$ (giving the conditional distribution $p(X_i \vert \textrm{pa}(X_{i})) = \mathcal{N}(\sum_{j} w_{ji}X_{j},0.1)$ as described in \Secref{sec:task}). %, and represents the specific randomness introduced when taking one sample from the \CBN. 
After observing the nodes $X_{2:N}$ ($X_{1}$ is hidden) in the \CBN~in one sample, we can infer this latent randomness $\epsilon_i$ for each observable node $X_i$ (\ie~\textit{abduction} as described in the Supplementary Material) and answer counterfactual questions like "What would the values of the nodes be, had $X_i$ instead taken on a different value than what we observed?", for any of the observable nodes $X_i$. We test three new agents, two of which are learned: "Active Counterfactual", "Random Counterfactual", and "Optimal Counterfactual Baseline" (not learned).

\textbf{Counterfactual Agents:}
This agent is exactly analogous to the Interventional agent, with the addition that the latent randomness in the last information phase step $t = T-1$ (where say some $X_p=+5$), is stored and the same randomness is used in the quiz phase step $t = T$ (where say some $X_f=-5$). While the question our agents have had to answer correctly so far in order to maximize their reward in the quiz phase was "Which of the nodes $X_{2:N}$ will have the highest value when $X_f$ is set to $-5$?", in this setting, we ask "Which of the nodes $X_{2:N}$ would have had the highest value in the last step of the information phase, if instead of having the intervention $X_p=+5$,  we had the intervention $X_f = -5$?". 
We run active and random versions of this agent as described in \Secref{sec:task}.

\textbf{Optimal Counterfactual Baseline:}
This baseline receives the true \CBN~and does exact abduction of the latent randomness based on observations from the penultimate step of the information phase, and combines this correctly with the appropriate interventional inference on the true \CBN~in the quiz phase.

\subsection*{Results}

We focus on two key questions in this experiment. 

(i) Do our agents learn to do counterfactual inference? The Active-Counterfactual Agent achieves higher reward than the Active-Interventional Agent ($p=$ \num{2e-5}). To evaluate whether this difference results from the agent's use of abduction (see the Supplementary Material for details), we split the test set into two groups, depending on whether or not the decision for which node will have the highest value in the quiz phase is affected by the latent randomness, i.e. whether or not the node with the maximum value in the quiz phase changes if the noise is resampled.
This is most prevalent in cases where the maximum expected reward is degenerate, i.e. where several nodes give the same maximum reward (denoted by hatched bars in Figure \ref{fig:expt3}b). Here, agents with no access to the randomness have no basis for choosing one over the other, but different noise samples can give rise to significant differences in the actual values that these degenerate nodes have. 
We see indeed that there is no difference in the rewards received by the Active-Counterfactual and Active-Interventional Agents in the cases where the maximum values are distinct, however the Active-Counterfactual Agent significantly outperforms the Active-Interventional Agent in cases where there are degenerate maximum values. 
\begin{wrapfigure}{l}{0.27\textwidth}
\centering
\vspace{-0.4cm}
\includegraphics[width=0.3\textwidth]{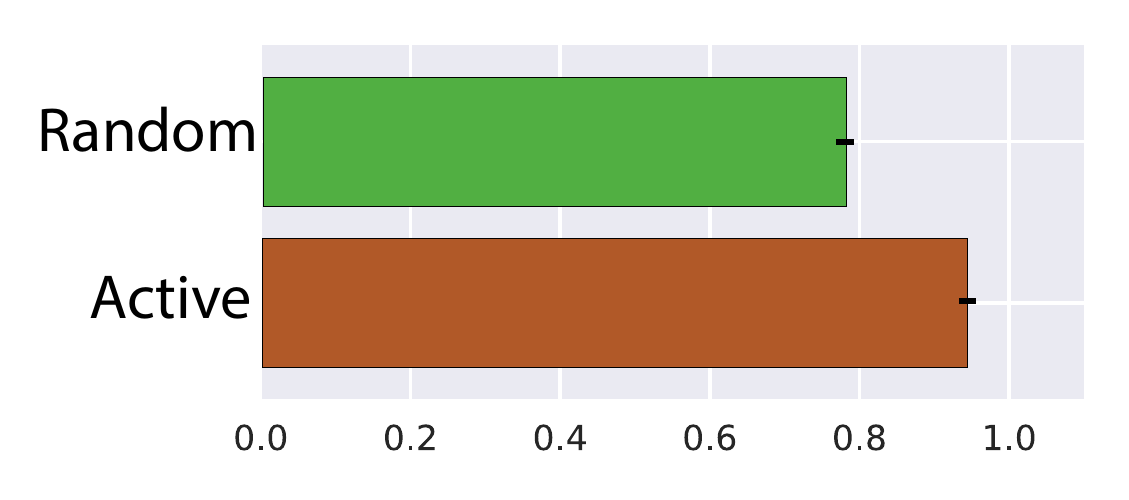} 
\vspace{-20pt}
\caption{Active and Random Counterfactual Agents}
\label{fig:expt3_active}
\vspace{-0.4cm}
\end{wrapfigure}(ii) Do our agents learn to make useful interventions in the service of a counterfactual task?  The Active-Counterfactual Agent's performance is significantly greater than the Random-Counterfactual Agent's (\Figref{fig:expt2_active}). This indicates that when the agent is allowed to choose its actions, it makes tailored, non-random choices about the interventions it makes and the data it wants to observe -- even in the service of a counterfactual objective. 

\section{Summary of Results}

In this paper we used the meta-learning to train a recurrent network -- using model-free reinforcement learning -- to implement an algorithm capable of causal reasoning. Agents trained in this manner performed causal reasoning in three data settings: observational, interventional, and counterfactual. Crucially, our approach did not require explicit encoding of formal principles of causal inference. Rather, by optimizing an agent to perform a task that depended on causal structure, the agent learned implicit strategies to generate and use different kinds of available data for causal reasoning, including drawing causal inferences from passive observation, actively intervening, and making counterfactual predictions, all on held out causal \CBNs~that the agents had never previously seen. 

A consistent result in all three data settings was that our agents learned to perform good experiment design or \textit{active learning}. That is, they learned a non-random data collection policy where they actively chose which nodes to intervene (or condition) on in the information phase, and thus could control the kinds of data they saw, leading to higher performance in the quiz phase than that from an agent with a random data collection policy. Below, we summarize the other keys results from each of the three experiments.

In Section 4.1 and \Figref{fig:expt1}, we showed that agents learned \textit{to perform do-calculus}. In \Figref{fig:expt1}a we saw that, the trained agent with access to only observational data received more reward than the highest possible reward achievable without causal knowledge. We further observed in \Figref{fig:expt1}b that this performance increase occurred selectively in cases where do-calculus made a prediction distinguishable from the predictions based on correlations -- \ie~where the externally intervened node had a parent, meaning that the intervention resulted in a different graph.

In Section 4.2 and \Figref{fig:expt2}, we showed that agents learned \textit{to resolve unobserved confounders using interventions} (which is impossible with only observational data). In \Figref{fig:expt2}b we saw that agents with access to interventional data performed better than agents with access to only observational data only in cases where the intervened node shared an unobserved parent (a confounder) with other variables in the graph.

In Section 4.3 and \Figref{fig:expt3}, we showed that agents learned \textit{to use counterfactuals}. In \Figref{fig:expt3}a we saw that agents with additional access to the specific randomness in the test phase performed better than agents with access to only interventional data. In \Figref{fig:expt3}b, we found that the increased performance was observed only in cases where the maximum mean value in the graph was degenerate, and optimal choice was affected by the latent randomness -- \ie~where multiple nodes had the same value on average and the specific randomness could be used to distinguish their actual values in that specific case.

\section{Discussion and Future Work}

To our knowledge, this is the first direct demonstration that causal reasoning can arise out of model-free reinforcement learning. Our paper lays the groundwork for a meta-reinforcement learning approach to causal reasoning that potentially offers several advantages over formal methods for causal inference in complex real world settings.
%Compared to hand-crafted causal reasoning algorithms, the meta-learning approach has several advantages.

First, traditional formal approaches usually decouple the problems of \emph{causal induction} (inferring the structure of the underlying model from data) and \emph{causal inference} (estimating causal effects based on a known model). Despite advances in both \citep{ortega2015causal,lattimore2016causal,bramley2017formalizing,forney2017counterfactual,sen2017identifying,parida2018multivariate}, inducing models is expensive and typically requires simplifying assumptions. When induction and inference are decoupled, the assumptions used at the induction step are not fully optimized for the inference that will be performed downstream. By contrast, our model learns to perform induction and inference end-to-end, and can potentially find representations of causal structure best tuned for the required causal inferences. Meta-learning can sometimes even leverage structure in the problem domain that may be too complex to specify when inducing a model \cite{duan2016RL2,santoro2016meta,wang2016}, allowing more efficient and accurate causal reasoning than would be possible without representing and exploiting this structure.

Second, since both the induction and inference steps are costly, formal methods can be very slow at run-time when faced with a new query. Meta-learning shifts most of the compute burden from inference time to training time. This is advantageous when training time is ample but fast answers are needed at run-time.

Finally, by using an RL framework, our agents can learn to take actions that produce useful information---i.e. perform active learning. Our agents' active intervention policy performed significantly better than a random intervention policy, which demonstrates the promise of learning an experimentation policy end-to-end with the causal reasoning built on the resulting observations.

Our work focused on a simple domain because our aim was to test in the most direct way possible whether causal reasoning can emerge from meta-learning. Follow-up work should focus on scaling up our approach to larger environments, with more complex causal structure and a more diverse range of tasks. This opens up possibilities for agents that perform active experiments to support structured exploration in RL, and learning optimal experiment design in complex domains where large numbers of random interventions are prohibitive. The results here are a first step in this direction, obtained using relatively standard deep RL components -- our approach will likely benefit from more advanced architectures \citep[e.g.][]{hester2017deep,hessel2018multi,espeholt2018impala} that allow us to train on longer more complex episodes, as well as models which are more explicitly compositional \citep[e.g.][]{andreas2016neural,battaglia2018relational} or have richer semantics \citep[e.g.][]{ganin2018synthesizing} that can more explicitly leverage symmetries in the environment and improve generalization and training efficiency.

\section{Acknowledgements}
The authors would like to thank the following people for helpful discussions and comments: Neil Rabinowitz, Neil Bramley, Tobias Gerstenberg, Andrea Tacchetti, Victor Bapst, Samuel Gershman.

% The authors would like to thank the following people for helpful discussions and comments: Misha Denil, Neil Rabinowitz, Ian Dunning, Adam Santoro.

% \bibliographystyle{iclr2019_conference}
\bibliographystyle{plainnat}
\bibliography{references}

\newpage

\end{document}

\maketitle

\section{Additional Baselines}

\begin{wrapfigure}{l}{0.27\textwidth}
\centering
\vspace{-0.5cm}
\includegraphics[width=0.3\textwidth]{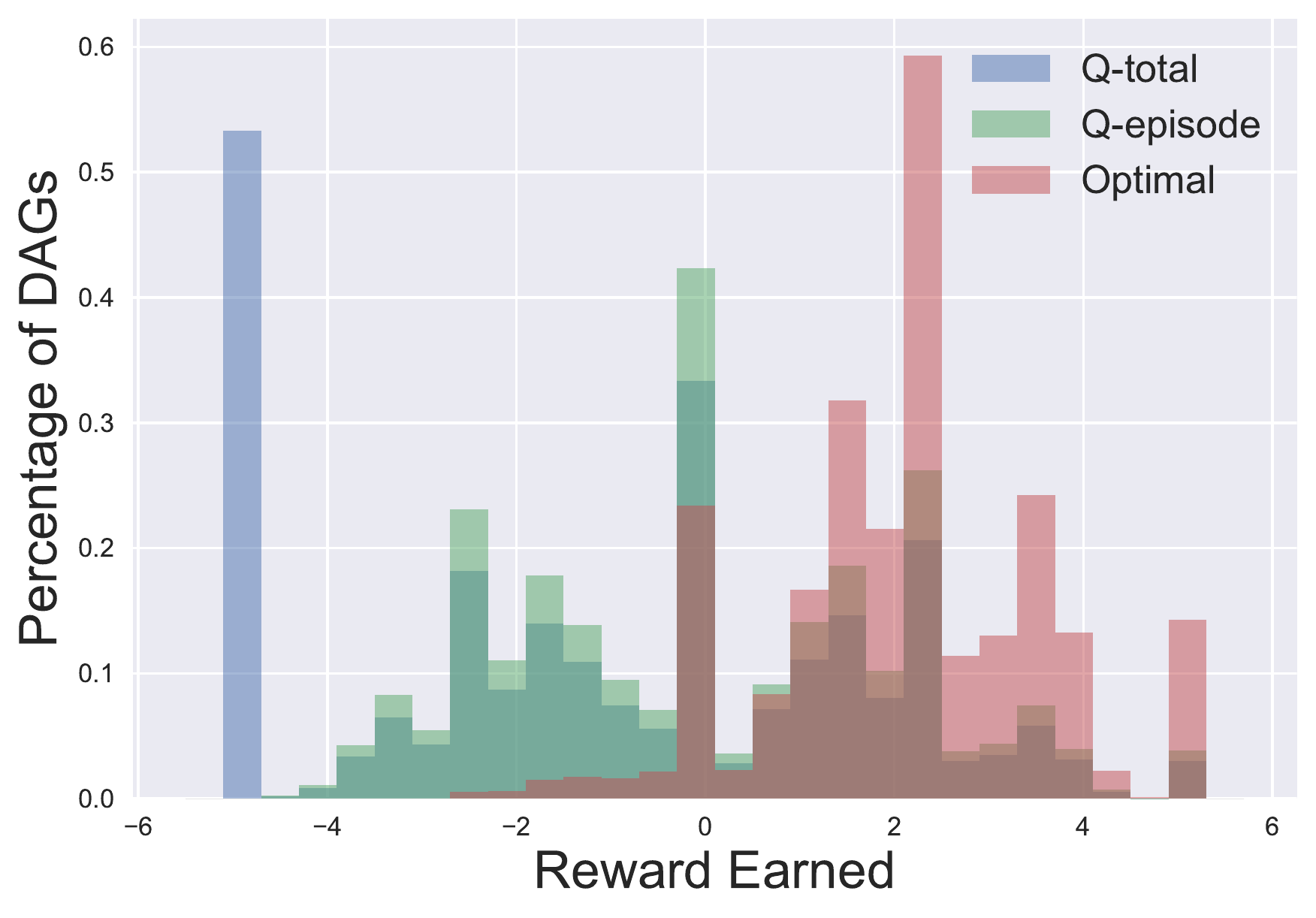} 
\vspace{-0.5cm}
\caption{Reward distribution}
\label{fig:rewarddist}
\vspace{-0.4cm}
\end{wrapfigure}

We can also compare the performance of these agents to two standard model-free RL baselines. The Q-total Agent learns a Q-value for each action across all steps for all the episodes. The Q-episode Agent learns a Q-value for each action conditioned on the input at each time step $[o_t, a_{t-1}, r_{t-1}]$, but with no LSTM memory to store previous actions and observations. Since the relationship between action and reward is random between episodes, Q-total was equivalent to selecting actions randomly, resulting in a considerably negative reward ($-1.247 \pm
2.940$). The Q-episode agent essentially makes sure to not choose the arm that is indicated by $m_t$ to be the external intervention (which is assured to be equal to $-5$), and essentially chooses randomly otherwise, giving a reward close to 0 ($0.080 \pm 2.077$).

\section{Formalism for Memory-based Meta-learning \label{sec:meta-formalism}}
Consider a distribution ${\cal D}$ over Markov Decision Processes (MDPs). We train an agent with memory (in our case an RNN-based agent) on this distribution. In each episode, we sample a task $m\sim {\cal D}$. At each step $t$ within an episode, the agent sees an observation $o_t$, executes an action $a_t$, and receives a reward $r_t$. Both $a_{t-1}$ and $r_{t-1}$ are given as additional inputs to the network. Thus, via the recurrence of the network, each action is a function of the entire trajectory ${\cal H}_{t}=\{ o_0, a_0, r_0, \dots, o_{t-1}, a_{t-1}, r_{t-1}, o_t\}$ of the episode. Because this function is parameterized by the neural network, its complexity is limited only by the size of the network.

\section{Abduction-Action-Prediction Method for Counterfactual Reasoning}

\citet{pearl16causal}'s ``abduction-action-prediction'' method prescribes one way to answer counterfactual queries of the type "What would the cardiac health of individual $i$ have been had she done more exercise?", by estimating the specific latent randomness in the unobserved makeup of the individual and by transferring it to the counterfactual world.
Assume, for example, the following model for ${\cal G}$ of Section 2.1%\Secref{main-sec:causality}
: $E=w_{AE}A + \eta$, $H=w_{AH}A+w_{EH}E+\epsilon$,
% where $\eta$ and $\epsilon$ are unobserved Gaussian representing the observed makeup. Suppose that individual $i$ has associated values $a^i, e^i, h^i$.
where the weights $w_{ij}$ represent the known causal effects in $\cal G$ and $\epsilon$ and $\eta$ are terms of (e.g.) Gaussian noise that represent the latent randomness in the makeup of each individual\footnote{These are zero in expectation, so without access to their value for an individual we simply use ${\cal G}$: $E=w_{AE}A$, $H=w_{AH}A+w_{EH}E$ to make causal predictions.}. Suppose that for individual $i$ we observe: $A = a^i$, $E = e^i$, $H = h^i$. We can answer the counterfactual question of "What if individual $i$ had done more exercise, \ie~$E=e'$, instead?" by: a) \emph{Abduction:} estimate the individual's specific makeup with $\epsilon^i=h^i-w_{AH}a^i-w_{EH}e^i$, b) \emph{Action:} set $E$ to more exercise $e'$, c) \emph{Prediction:} predict a new value for cardiac health as $h'=w_{AH}a^i+w_{EH}e'+\epsilon^i$.  

\section{Additional Experiments}

The purview of the previous experiments was to show a proof of concept on a simple tractable system, demonstrating that causal induction and inference can be learned and implemented via a meta-learned agent. In the following, we scale up to more complex systems in two new experiments.

\subsection{Experiment 4: Non-linear Causal Graphs}

\begin{figure}[ht!]
\centering
\includegraphics[width=0.46\textwidth]{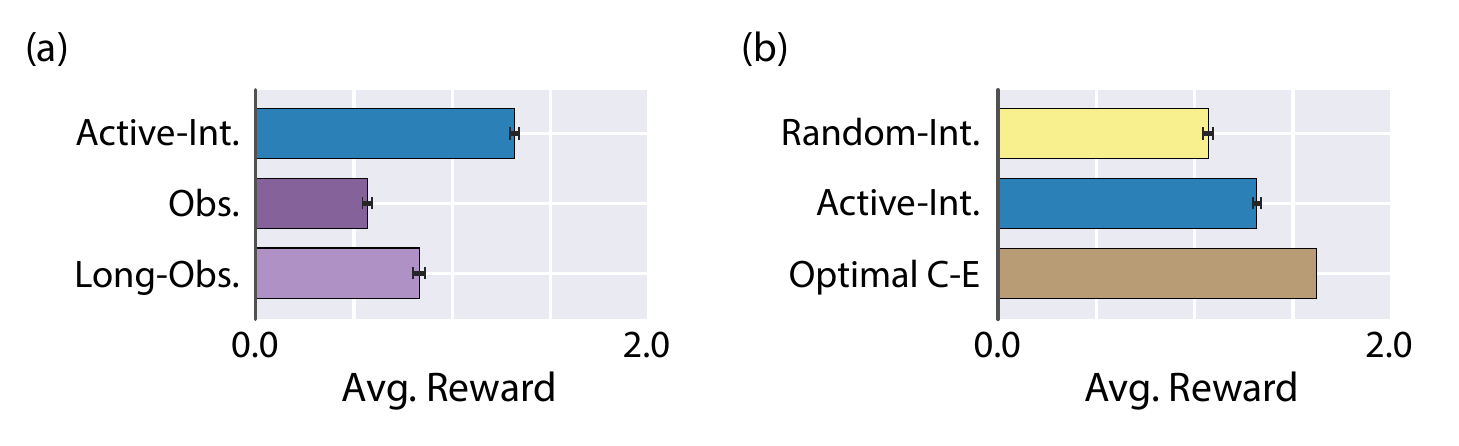} 
\caption{Results for non-linear graphs. (a) Comparing agent performances with different data. (b) Comparing information phase intervention policies. }
\label{fig:nonlinear_results}
\vspace{-0.3cm}
\end{figure}

In this experiment, we generalize some of our results to nonlinear, non-Gaussian causal graphs which are more typical of real-world causal graphs and to demonstrate that our results hold without loss of generality on such systems. 

Here we investigate causal Bayesian networks (\CBNs) with a quadratic dependence on the parents by changing the conditional distribution to $p(X_{i}|\textrm{pa}(X_{i})) = \mathcal{N}(\frac{1}{N_{i}} \sum_{j} w_{ji}(X_{j} + X_{j}^2), \sigma)$. Here, although each node is normally distributed given its parents, the joint distribution is not multivariate Gaussian due to the non-linearity in how the means are determined. We find that the Long-Observational Agent achieves more reward than the Observational Agent indicating that the agent is in fact learning the statistical dependencies between the nodes, within an episode\footnote{The conditional distribution $p(X_{1:N\setminus j }|X_j=5)$, and therefore Conditional Agents, were non-trivial to calculate for the quadratic case.}. We also find that the Active-Interventional Agent achieves reward well above the best agent with access to only observational data (Long-Observational in this case) indicating an ability to reason from interventions. We also see that the Active-Interventional Agent performs better than the Random-Interventional Agent indicating an ability to choose informative interventions.

\subsection{Experiment 5: Larger Causal Graphs}

\begin{figure}[ht!]
\centering
\includegraphics[width=0.46\textwidth]{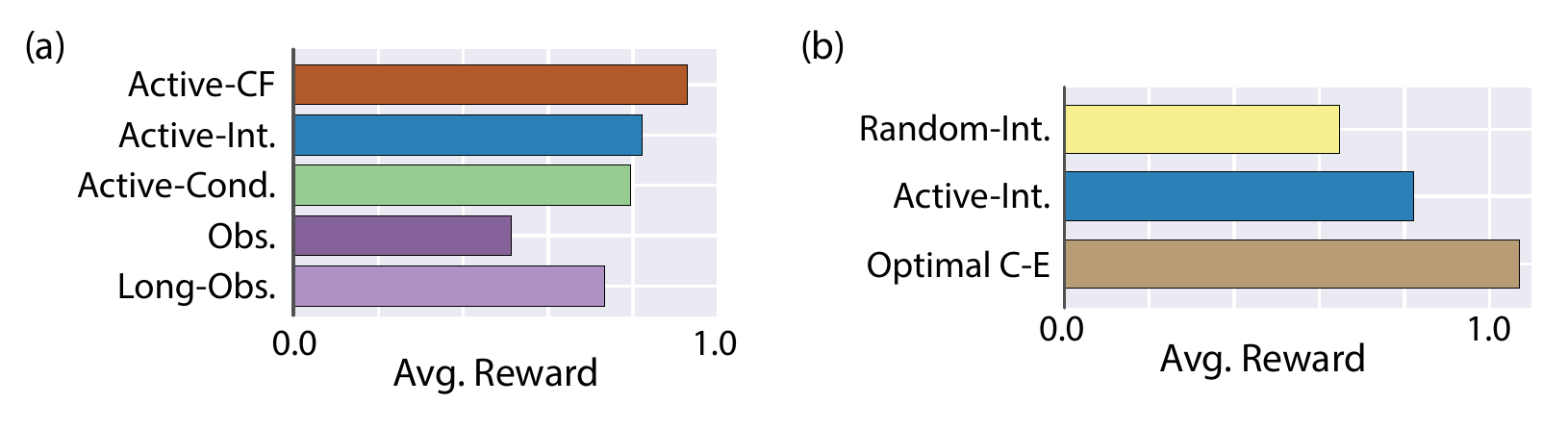} 
\caption{Results for $N=6$ graphs. (a) Comparing agent performances with different data. (b) Comparing information phase intervention policies. }
\label{fig:N6_results}
\vspace{-0.3cm}
\end{figure}
In this experiment we scaled up to larger graphs with $N=6$ nodes, which afforded considerably more unique \CBNs~than with $N=5$ ($1.4\times10^7$ vs $5.9\times10^4$). As shown in \Figref{fig:N6_results}a, we find the same pattern of behavior noted in the main text where the rewards earned are ordered such that Observational agent $<$ Active-Conditional agent $<$ Active-Interventional agent $<$ Active-Counterfactual agent. We see additionally in \Figref{fig:N6_results}b that the Active-Interventional agent performs significantly better than the baseline Random-Interventional agent, indicating an ability to choose non-random, informative interventions.

\section{Causal Bayesian Networks\label{sec:BN}}
By combining graph theory and probability theory, the causal Bayesian network framework provides us with a graphical tool to formalize and test different levels of causal reasoning. This section introduces the main definitions underlying this framework and explains how to visually test for statistical independence  \citep{pearl88probabilistic,bishop06pattern,kollerl09probabilistic,barber12bayesian,murphy12machine}.

\begin{figure}[t]
% \vspace{-0.95cm}
\begin{center}
\subfloat[]{\scalebox{0.7}{
\begin{tikzpicture}[dgraph]
\node[ocont] (x2) at (4.5,0) {$X_2$};
\node[ocont] (x1) at (1.5,0) {$X_1$};
\node[ocont] (x3) at (3,0) {$X_3$};
\node[ocont] (x4) at (3,-1.5) {$X_4$};
\draw[line width=1.15pt](x1)--(x3);\draw[line width=1.15pt](x2)--(x3);\draw[line width=1.15pt](x3)--(x4);\draw[line width=1.15pt](x2)to [bend left=35](x4);
\end{tikzpicture}}}
\hskip0.2cm
\subfloat[]{\scalebox{0.7}{
\begin{tikzpicture}[dgraph]
\node[ocont] (x2) at (4.5,0) {$X_2$};
\node[ocont] (x1) at (1.5,0) {$X_1$};
\node[ocont] (x3) at (3,0) {$X_3$};
\node[ocont] (x4) at (3,-1.5) {$X_4$};
\draw[line width=1.15pt](x1)--(x3);\draw[line width=1.15pt](x2)--(x3);\draw[line width=1.15pt](x3)--(x4);\draw[line width=1.15pt](x2)to [bend left=35](x4);\draw[line width=1.15pt](x4)to [bend left=35](x1);
\end{tikzpicture}}}
\vspace{-0.15cm}
\caption{(a): Directed acyclic graph. The node $X_3$ is a collider on the path $X_1 \rightarrow X_3 \leftarrow X_2$ and a non-collider on the path $X_2\rightarrow X_3\rightarrow X_4$. (b): Cyclic graph obtained from (a) by adding a link from $X_4$ to $X_1$.}
\label{fig:indep_HMM}
\vspace{-0.75cm}
\end{center}
% \vskip-0.3cm
\end{figure}
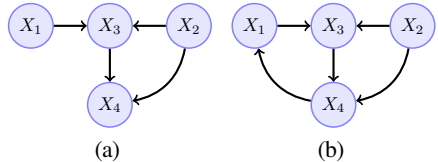

A {\bf graph} is a collection of nodes and links connecting pairs of nodes.
The links may be directed or undirected, giving rise to {\bf directed} or {\bf undirected graphs} respectively.\\[5pt]
A {\bf path} from node $X_i$ to node $X_j$ is a sequence of linked nodes starting
at $X_i$ and ending at $X_j$. A {\bf directed path} is a path whose links are directed and pointing from preceding towards following nodes in the sequence.\\[5pt]
%\jane{add def CBN here}
A {\bf directed acyclic graph} is a directed graph with no
directed paths starting and ending at the same node. For example, the directed graph in \Figref{fig:indep_HMM}(a) is acyclic. 
The addition of a link from $X_4$ to $X_1$ gives rise to a cyclic graph (\Figref{fig:indep_HMM}(b)).\\[5pt]
A node $X_i$ with a directed link to $X_j$ is called {\bf parent}
of $X_j$. In this case, $X_j$ is called {\bf child} of $X_i$.\\[5pt]
A node is a {\bf collider} on a specified path if it has (at least) two parents on that path.
Notice that a node can be a collider on a path and a non-collider on another path. For example, in \Figref{fig:indep_HMM}(a) $X_3$ is a collider on the path $X_1 \rightarrow X_3 \leftarrow X_2$ and a non-collider on the path $X_2\rightarrow X_3\rightarrow X_4$.\\[5pt]
A node $X_i$ is an {\bf ancestor} of a node $X_j$ if there exists a directed path from $X_i$ to $X_j$. In this case, $X_j$ is a {\bf descendant} of $X_i$.\\[5pt]
A {\bf graphical model} is a graph in which nodes represent random variables and links express statistical relationships between the variables.\\[5pt]
A {\bf Bayesian network} is a directed acyclic graphical model in which each node $X_i$ is associated
with the conditional distribution $p(X_i|\text{pa}(X_i))$, where $\text{pa}(X_i)$ indicates the parents of $X_i$. The joint distribution of all nodes in
the graph, $p(X_{1:N})$, is given by the product of all conditional distributions, i.e. $p(X_{1:N})=\prod_{i=1}^Np(X_i|\text{pa}(X_i))$.

When equipped with causal semantic, namely when describing the process underlying the data generation, a Bayesian network expresses both causal and statistical relationships among random variables---in such a case the network is called {\bf causal}.  

\paragraph{Assessing statistical independence in Bayesian networks.}
Given the sets of random variables ${\cal X}, {\cal Y}$ and ${\cal Z}$,
${\cal X}$ and ${\cal Y}$ are statistically independent given ${\cal Z}$ if all paths
from any element of ${\cal X}$ to any element of ${\cal Y}$ are {\bf closed} (or {\bf blocked}). A path is closed if at least one of the following conditions is satisfied:
\begin{enumerate}
\setlength{\itemsep}{-2pt}  
\setlength{\parsep}{-4pt}
\item[(i)] There is a non-collider on the path which belongs to the conditioning set ${\cal Z}$.
\item[(ii)] There is a collider on the path such that neither the collider nor any of its descendants belong to ${\cal Z}$.
\end{enumerate}

\bibliographystyle{plainnat}
\bibliography{references}